\documentclass[letterpaper, conference, 10pt]{style/ieeeconf}
\usepackage[T1]{fontenc}
\usepackage[latin9]{inputenc}
\usepackage{calc}
\usepackage{amssymb, amsmath}
\usepackage{graphicx}
\usepackage{xcolor}
\usepackage{mathtools}
\usepackage{cite}
\usepackage{hyperref}
\usepackage[capitalize]{cleveref}
\usepackage{subfigure}
\usepackage[font=footnotesize, skip=1ex, belowskip=-10pt]{caption}
\usepackage{multirow}
\usepackage{algorithm}
\usepackage{algpseudocode}
\usepackage[percent]{overpic}
\usepackage{scalerel}
\usepackage{float}

\DeclareMathOperator*{\argmin}{arg\,min}

\begin{document}
\title{Incremental cycle bases for cycle-based pose graph optimization}

\author{Brendon Forsgren$^1$, Kevin Brink$^2$, Prashant Ganesh$^3$, Timothy W. McLain$^1$

\thanks{$^1$ Brendon Forsgren and Timothy McLain are with the Department of Mechanical Engineering, Brigham Young University {\tt\footnotesize bforsgren29@gmail.com, mclaiin@byu.edu}}
\thanks{$^2$ Kevin Brink is with the Air Force Research Laboratory, Eglin Air Force Base {\tt\footnotesize kevin.brink@us.af.mil}}
\thanks{$^3$ Prashant Ganesh is with the Department of Mechanical and Aerospace Engineering, University of Florida {\tt\footnotesize prashant.ganesh@ufl.edu}}
}

\maketitle

\begin{abstract}
Pose graph optimization is a special case of the simultaneous localization
and mapping problem where the only variables to be estimated are pose
variables and the only measurements are inter-pose constraints. The vast
majority of pose graph optimization techniques are vertex based (variables are robot poses), but
recent work has parameterized the pose graph optimization problem in a relative
fashion (variables are the transformations between poses) that utilizes a
minimum cycle basis to maximize the sparsity of the problem. We explore the
construction of a cycle basis in an incremental manner while maximizing the
sparsity. We validate an algorithm that constructs a sparse cycle basis
incrementally and compare its performance with a minimum cycle basis.
Additionally, we present an algorithm to approximate the minimum cycle
basis of two graphs that are sparsely connected as is common in multi-agent
scenarios. Lastly, the relative parameterization of pose graph optimization has
been limited to  using rigid body transforms on $SE(2)$ or $SE(3)$ as the
constraints between poses. We introduce a methodology to allow for the
use of lower-degree-of-freedom measurements in the
relative pose graph optimization problem. We provide extensive validation of
our algorithms on standard benchmarks, simulated datasets, and custom
hardware.
\end{abstract}

\section{Introduction}

The simultaneous localization and mapping (SLAM) is
often modeled as a factor graph with variable
nodes consisting of both pose and landmark variables, and factor nodes
representing constraints between variables. A special case of SLAM, called
pose graph optimization (PGO), occurs when the only variable nodes are robot
poses and the only factor nodes are constraints between poses.

In robotics, PGO has become a common technique to estimate the discrete-time
trajectory of a robot when only the robot trajectory is of interest, as in
the cases described in \cite{wheeler2018relative,ellingson2020cooperative}.
PGO is usually solved using nonlinear least squares (NLS) optimization
techniques and, as such, requires an initial estimate to start the optimization.
The convergence of the problem to the global minima requires that the initial
guess be in the basin of convergence of the global minima or otherwise a
suboptimal local minima will be found.

\subsection{Related Work}

The birth of modern SLAM algorithms started with the seminal paper by Lu and
Milos \cite{lu1997globally}. Since then, the robotics community has developed
many fast and robust algorithms to enable autonomous agents to solve the SLAM
problem. Various techniques have been developed to increase the robustness of
SLAM algorithms including obtaining a good initialization
\cite{carlone2014angular, carlone2015initialization}, using robust cost
functions \cite{agarwal2013robust,aloise2019chordal}, using convex relaxations
\cite{liu2012convex,rosen2015convex,rosen2020scalable }, and finding globally
optimal solutions \cite{rosen2019se, mangelson2019guaranteed}.
At their core, these algorithms exploit the sparse nature of SLAM to
efficiently compute estimates of the variables using sparse nonlinear
solvers such as GTSAM \cite{dellaert2012factor} or g2o \cite{grisetti2011g2o}.
The sparsity of the problem is maintained by methods such as variable
reordering \cite{agarwal2012variable}. The iSAM2 solver presented in
\cite{kaess2011isam2}, in addition to doing online variable reordering,
provides a method to incrementally update the solution by only updating the
affected variables.

Recent work has focused on a relative parameterization of PGO
\cite{bai2018robust, jackson2019direct, bai2021sparse} where each robot pose is
expressed in the frame of the pose that precedes it. Olson et al.
\cite{teller2006fast} notes that the relative parameterization of PGO loses the
sparse property present in the traditional global parameterization.
Jackson et al. \cite{jackson2019direct} show that the relative
parameterization is better conditioned than the traditional global
parameterization. Bai et al. \cite{bai2018robust} reformulate the problem as a
constrained optimization problem by enforcing that cycles in the graph return
to the origin when traversed. Bai notes
that the set of cycles must form a cycle basis for the graph being optimized.
In his most recent work \cite{bai2021sparse}, Bai presents a method to maximize
the sparsity of the relative parameterization by forcing the set of cycle
constraints to form a minimum cycle basis (MCB) and
shows that the run time of the relative parameterization is comparable to that
of the global parameterization. The primary drawback to the method is that
computing a MCB is an expensive procedure, showcasing the need for methods to
incrementally update the MCB.

Constructing a MCB is a well studied process in graph theory with the first
polynomial time algorithm being developed by Horton
\cite{horton1987polynomial}. The MCB can be found by first finding the all
pairs shortest paths (APSP) data structure and then creating a Horton set. The
Horton set is a set of cycles known to be a superset of a MCB. The set
is ordered by weight and  the MCB is found by taking the shortest cycles
that form a cycle basis for the graph. Other algorithms to construct or
approximate a MCB have been proposed
\cite{amaldi2009breaking,kavitha2011new,mehlhorn2009minimum,rathod2021fast}
that are less computationally complex than Horton's algorithm. However, all of
these methods are for static graphs, or graphs without edge insertions or
deletions.
To our knowledge, no incremental algorithm to construct a MCB exists.
Bai describes an algorithm in \cite{bai2018robust} based on a fundamental cycle
basis that uses the odometry backbone to generate a sparse fundamental cycle
basis. In his dissertation \cite{bai2020two}, Bai describes a
greedy algorithm to incrementally approximate a MCB but does not
evaluate the algorithm in the dissertation or any peer reviewed work.

The contributions made in this paper are as follows:
\begin{enumerate}
    \item Extensive validation of the greedy incremental algorithm presented in \cite{bai2020two}.
    \item An algorithm to generate a cycle basis of multiple graphs that are sparsely connected.
    \item A methodology to use low-degree-of-freedom measurements in the relative parameterization of PGO.
    \item Validation of the proposed algorithms on standard benchmarks, simulated data, and in hardware experiments.
\end{enumerate}

\section{Problem Definition}

Here we define the PGO problem to be solved, relevant technical
definitions, and other notation. A graph $\mathcal{G}$ is defined as
$\mathcal{G} = \mathcal{G}(\mathcal{V}, \mathcal{E})$ where $\mathcal{G}$ is
the graph, $\mathcal{V}$ are vertices, and $\mathcal{E}$ are the edges. A path,
$\mathcal{P}$, in $\mathcal{G}$ is a subgraph of $\mathcal{G}$ where
all vertices have degree two except for two vertices, which have degree one. A
cycle in $\mathcal{G}$, denoted as $\mathcal{C}$, is
a subgraph of $\mathcal{G}$ where all vertices have an even degree.
The cycle space of $\mathcal{G}$ is the set of all the cycles in $\mathcal{G}$
and a cycle basis of $\mathcal{G}$, denoted as $\mathcal{B}$, is a set of
independent cycles from which any
cycle in $\mathcal{G}$ can be created by combining cycles in $\mathcal{B}$.
It is known that the dimension of the
cycle basis is $\nu = \lvert \mathcal{E} \rvert - \lvert \mathcal{V} \rvert + 1$ \cite{bai2021sparse},
where $\lvert \cdot \rvert$ denotes the cardinality of the set.

Now we define the PGO problem that we wish to solve. Given a graph
$\mathcal{G} = \mathcal{G}(\mathcal{V, E})$, where $\mathcal{V}$ are robot poses
and $\mathcal{E}$ are the rigid body transformations (in 2D or 3D) between
poses, we wish to estimate the edges in $\mathcal{E}$.
The estimates are found by solving the following problem defined in
\cite{bai2021sparse}

\begin{equation}
\begin{aligned}
    \{\mathbf{T}_k\}_{k \in \mathcal{E}} &= \argmin \sum_{k \in \mathcal{E}} \lVert \mathbf{Log}(\mathbf{\tilde{T}}_k^{-1}\mathbf{T}_k) \rVert _{\Sigma_k} ^2 \\
    &\textbf{s.t.} \;\;  \mathbf{I} = \prod_{\mathbf{T}_k \in \mathcal{C}_i} \mathbf{T}_k \;\; \forall \;\; \mathcal{C}_i \in \mathcal{B}
\end{aligned}
\label{eq:cb_pgo}
\end{equation}
where $\mathbf{T}_k$ is an edge in $\mathcal{E}$ to be estimated, $\mathbf{\tilde{T}}_k$ is the
measurement of edge $\mathbf{T}_k$, and $\mathcal{C}$ is a cycle in
$\mathcal{B}$. From this solution the robot poses can be calculated by
composing the rigid body transformations from the origin to the desired vertex
along any path in $\mathcal{G}$.

\section{Methods}
\label{sec:Methods}

Given the incremental nature of PGO problems in robotics, being able to
incrementally create a sparse cycle basis to constrain the problem defined in
\cref{eq:cb_pgo} is desirable. However, it is known that a MCB does not behave
well under standard graph operations (insertions, deletions, ect.),
\cite{zare2011cycle} meaning that there is no known way to update the MCB of a
graph under a given operation. In this section we provide an overview of the
incremental algorithm to construct a cycle basis presented in
\cite{bai2020two}. We additionally present an algorithm that will
incrementally approximate the MCB of two disjoint graphs that become connected
as is the case in multi-agent scenarios where inter-vehicle measurements are
obtained. Lastly, we present a framework that will enable low degree of freedom
measurements to be used in the cycle-based PGO problem defined in
\cref{eq:cb_pgo}.

\subsection{Incremental Cycle Basis Algorithm}

Bai, in his dissertation \cite{bai2020two}, presents the following algorithm to
incrementally approximate a MCB. Assume that a previously valid cycle basis,
$\mathcal{B}_k$ for a graph $\mathcal{G}$ exists and that a new edge $e_{ij}$
is introduced into the graph where $i$ and $j$ are non-sequential nodes. An
approximate MCB of $\mathcal{G}$ can be defined as
$\mathcal{B}_{k+1} = \mathcal{B}_k \cup \mathcal{C}_{ij}$ where
$\mathcal{C}_{ij}$ is any cycle that contains $e_{ij}$. The sparsity of
$\mathcal{B}_{k+1}$ can be maximized by defining the cycle $\mathcal{C}_{ij}$
such that it has minimum length. This is accomplished by letting
$\mathcal{C}_{ij} = \mathcal{P}(i, j) \cup e_{ij}$ where $\mathcal{P}(i, j)$ is
the shortest path between nodes $i$ and $j$. A proof that this algorithm forms
a valid cycle basis can be found in \cite{bai2020two}.


The complexity of this algorithm is
much lower than that of the MCB algorithm presented in
\cite{bai2021sparse}. Bai notes that the most complex portion of his MCB
algorithm is constructing an APSP structure and has a complexity of
$O(nm\text{log}n + n^2 \text{log}n + nm)$ where $n=|\mathcal{V}|$ and
$m=|\mathcal{E}|$. In comparison, the incremental algorithm
only requires finding the shortest path between two nodes, which can be done
with a breadth first search and a complexity of $O(n + m)$.

\subsection{Multi-agent Incremental Cycle Basis Algorithm}

In this section we propose an algorithm that will approximate the MCB of a
graph that is created in multi-agent scenarios. Assume two robots are
collecting data and each are forming
their own graphs $\mathcal{G}^i$, where the superscript
denotes the robot ID. Each robot is maintaining and updating its own cycle
basis, be it an approximation or an exact MCB, $\mathcal{B}_k^i$, where the subscript $k$ denotes at timestep $k$.
The robots communicate periodically to share data. The data is used to form the
graph $\mathcal{G}^{ij}$ and to estimate any inter-robot loop closures.
The question becomes can
the MCB, $\mathcal{B}_k^{ij}$, be approximated, where the superscript denotes the
union of the two graphs and the connecting edges.

We begin by proving that a cycle basis can be constructed incrementally and
then outline a heuristic to decrease the computational complexity. We define
the set of inter-agent relative pose measurements $\mathcal{E}^{ab}_i = (e_0, e_1, ..., e_i)$,
and $\mathcal{P}(a_i, a_j)$ as the shortest path between nodes $i$ and $j$ on
agent $a$ and likewise for agent $b$. Assume that cycle bases $\mathcal{B}^a$
and $\mathcal{B}^b$ are valid when the second inter-vehicle measurement is
taken and $\mathcal{E}^{ab}_1$ is formed. It is easy to identify that
$\mathcal{B}^a \cup \mathcal{B}^b \cup \mathcal{B}^{ab}_1$, where
$\mathcal{B}^{ab}_1 = e_0 \cup \mathcal{P}(a_0, a_1) \cup e_1 \cup \mathcal{P}(b_0, b_1)$,
is a valid cycle basis since edges $e_0$ and $e_1$ are in no cycles in either
$\mathcal{B}^a$ or $\mathcal{B}^b$. Assuming that cycle bases $\mathcal{B}^a$
and $\mathcal{B}^b$ stay valid when the third inter-vehicle measurement is
taken then, cycle basis $\mathcal{B}^{ab}_2 = \mathcal{B}^{ab}_1 \cup
\mathcal{C}(e_2, \mathcal{E}^{ab}_1)$ is valid where $\mathcal{C}(e_2, \mathcal{E}^{ab}_1)$ is a
cycle containing edges $e_2$ and any edge in $\mathcal{E}^{ab}_1$. This process
can be continued to show that the cycle basis at timestep $k$ is valid so long
as the new cycle added to the basis contains edges $e_k$ and any edge in
$\mathcal{E}^{ab}_{k-1}$.

Having shown that we can incrementally form a cycle basis of the union of
connection of two graphs we note that the shortest cycle containing the new
edge, $e_k$ can be created by identifying which edge in $\mathcal{E}^{ab}_{k-1}$
makes the shortest cycle. This becomes computationally complex as the number of
edges in $\mathcal{E}^{ab}_{k-1}$ increases. We present our algorithm in
\cref{alg:multi_agent_cb} where we make the design decision to always use edges
$e_k$ and $e_{k-1}$ when constructing loops to cut down on the computational
complexity. Our algorithm is presented for the general case where multiple
inter-vehicle relative pose measurements may be detected at one time as in the case when
data is shared after a long period with no communication.

We note that our choice to use edges $e_k$ and $e_{k-1}$ when forming a loop
comes with no performance guarantees and that this choice of ordering in
\cref{alg:multi_agent_cb} will potentially produce longer cycles than another
heuristic. While a more optimal solution would be to select the edge in
$\mathcal{E}^{ab}_{k-1}$
that results in the shortest cycle, this becomes computationally complex as the
number of edges in $\mathcal{E}^{ab}$ increases since a large number of
shortest-path operations will need to be computed. Additional orderings of the
newly arrived edges could be used or the whole cycle basis could be recomputed
using an arbitrary order of the edges in $\mathcal{E}^{ab}_k$. The order of the
edges would surely impact the lengths of the produced cycles but determining an
optimal or near optimal ordering in an efficient manner is not obvious and is
outside the scope of this paper and thus left as a topic of future work.

We note that the algorithm in \cite{bai2020two} would also work for
constructing a cycle basis of
the the union of graphs $\mathcal{G}^i$ and $\mathcal{G}^j$. However, there are
several reasons why our algorithm would be preferable. The first is that our
algorithm is more scalable. Bai's algorithm operates over a single graph that
will be growing faster than the individual graphs of each agent, meaning that
graph operations such as BFS will take longer. The second benefit
of our algorithm is that ours is parallelizable. Since our algorithm consists of
operations over disconnected graphs, operations on different graphs can be done
in parallel. Constructing cycles in parallel also improves the scalability of
the algorithm by allowing more cycles to be constructed in the same processing
time.

Additionally, our algorithm is beneficial if there is sufficient
communication bandwidth for the agents to share their respective cycle bases.
The communication of the cycle bases $\mathcal{B}_k^i$ and $\mathcal{B}_k^j$
eliminate the need for these cycle bases to be recomputed. If such
communication is possible,
then the presented algorithm will eliminate the duplication of effort and the
only cycles to be computed are those that exist between agents. Lastly,
we note that this algorithm is also incremental and allows for the cycle basis
$\mathcal{B}^{ij}$ to be constructed as the inter-vehicle measurements arrive,
resulting in fast update times due to the low complexity of shortest path operations.

\begin{figure}
    \vspace{2mm}
    \centering
    \begin{overpic} [scale=0.5]{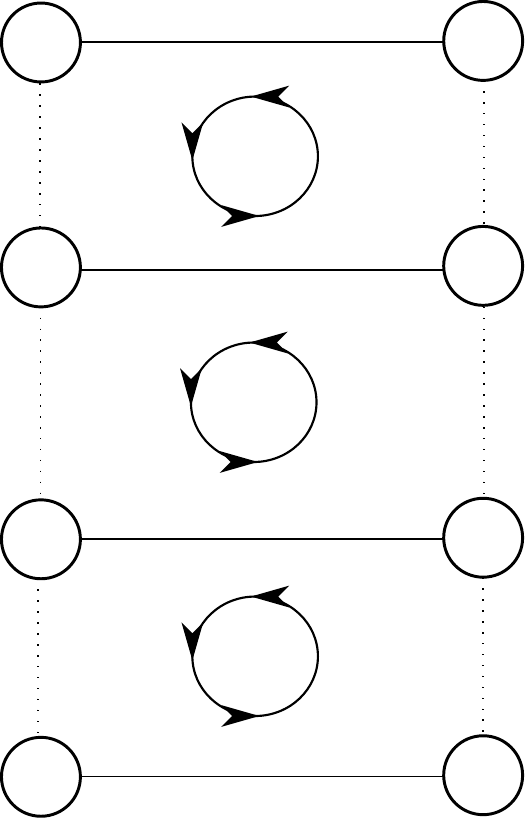}
        \put(2.5, 4){$a_{\scaleto{k}{4pt}}$}
        \put(0.75, 33){$a_{\scaleto{k+1}{4pt}}$}
        \put(0.75, 66.5){$a_{\scaleto{k+2}{4pt}}$}
        \put(0.75, 94){$a_{\scaleto{k+3}{4pt}}$}

        \put(57, 4){$b_{\scaleto{k}{4pt}}$}
        \put(55.5, 33){$b_{\scaleto{k+1}{4pt}}$}
        \put(55.5, 66.5){$b_{\scaleto{k+2}{4pt}}$}
        \put(55.5, 94){$b_{\scaleto{k+3}{4pt}}$}

        \put(28, 18){\large$\mathcal{C}_k$}
        \put(25.5, 49){\large$\mathcal{C}_{k+1}$}
        \put(25, 80){\large$\mathcal{C}_{k+2}$}

        \put(28, 2){\large$e_k$}
        \put(26.5, 31){\large$e_{k+1}$}
        \put(26.5, 64){\large$e_{k+2}$}
        \put(26.5, 92){\large$e_{k+3}$}

        \put(5, 17){$\mathcal{P}^i_k$}
        \put(5, 50){$\mathcal{P}^i_{k+1}$}
        \put(5, 80){$\mathcal{P}^i_{k+2}$}
        \put(52, 17){$\mathcal{P}^j_k$}
        \put(48, 50){$\mathcal{P}^j_{k+1}$}
        \put(48, 80){$\mathcal{P}^j_{k+2}$}

    \end{overpic}
    \caption{A visual description of the cycles created in \cref{alg:multi_agent_cb}. Dotted lines denote the shortest path between two nodes while solid lines denote the edge connecting two nodes.}
    \label{fig:ma_cycle_basis}
\end{figure}
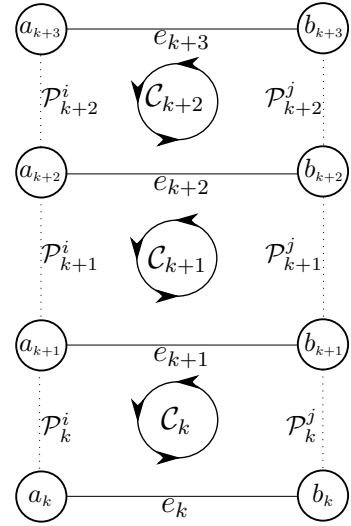


\begin{algorithm}
    \caption{Algorithm for finding a cycle basis incrementally for the connection of two graphs. \newline \textbf{INPUT:} $\mathcal{G}_i$, $\mathcal{G}_j$, $\mathcal{B}^{ij}_{k-1}$, $\mathcal{E}^{ij}_{k-1}$, $\mathcal{E}$. \newline \textbf{OUTPUT:} Updated Cycle Basis $\mathcal{B}^{ij}_k$.}
    \label{alg:multi_agent_cb}
\begin{algorithmic}
    \Function{ConnectedCycleBasis}{$\mathcal{G}_i$, $\mathcal{G}_j$, $\mathcal{E}^{ij}_{k-1}$, $\mathcal{E}$}
        \State $\mathcal{B}^{ij}_k \gets \mathcal{B}^{ij}_{k-1}$
        \For{$e_k$ in $\mathcal{E}$}
        \State $e_{k-1} \gets \mathcal{E}^{ij}_{k-1}[k-1]$
        \State $\mathcal{P}_i \gets$ Shortest path between nodes $a_k$ and $a_{k-1}$
        \State $\mathcal{P}_j \gets$ Shortest path between nodes $b_k$ and $b_{k-1}$
        \State $\mathcal{C} \gets \mathcal{P}_i \cup e_{k-1} \cup \mathcal{P}_j \cup e_k$
        \State $\mathcal{B}^{ij}_{k} \gets \mathcal{B}^{ij}_{k} \cup \mathcal{C}$
        \State $\mathcal{E}^{ij}_{k-1} \gets \mathcal{E}^{ij}_{k-1} \cup e_k$
        \EndFor
        \State $\mathcal{E}^{ij}_k \gets \mathcal{E}^{ij}_{k-1}$
    \EndFunction
\end{algorithmic}
\end{algorithm}

\subsection{Low-Degree-of-Freedom Measurements}

One of the shortcomings of the relative parameterization of PGO is that it
requires that the measurements between poses be a relative pose.
This requirement arises because of the cycle constraints used in the
optimization. However, measuring a relative pose between two poses can be
difficult, especially in multi-agent scenarios, and often low-degree-of-freedom (DOF)
measurements (range, bearing, etc.) are easier to obtain in real world scenarios. The cycle constraints
currently used to constrain the relative PGO problem are not well suited to
such measurements

If low-DOF were to be used there would be several
instances where it becomes unclear how to traverse a cycle. This is
especially true in multi-agent scenarios when there is no single odometry
backbone and two inter-agent measurements are required to form a cycle.
For example, there is no intuitive way to traverse a cycle that contains two
range measurements between different vehicles because there is not enough information
present to do so. In such scenarios, the cycle constraints lose their
usefulness, despite the fact that cycles still exist in the graph and the graph
structure is unchanged. 

Using low-DOF measurements requires a
different approach. One possible approach is to use the low-DOF
measurements to estimate a relative pose as done in
\cite{trawny2010interrobot,zhou2012determining} and then use the relative pose
in a cycle constraint. Another possible approach would be to
formulate a different kind of constraint that would use $n$
measurements, where $n$ measurements would be sufficient to uniquely satisfy the
constraint. While the number of measurements $n$ is known for many
measurement types, however, there is no concept or tool in graph theory, to our
knowledge, that can aid in the generation of such a constraint. As such, we
adopt the method of condensing low-DOF measurements into a single
rigid-body transformation.

The remainder of this section will apply this method using inter-vehicle range
and bearing measurements.

\subsubsection{Range and Bearing}

Estimating a relative pose from low-DOF measurements is a topic
that has received some attention in the robotics community.
In \cite{zhou2012determining} Zhou shows
that in three dimensions, the rotation between the two poses is unobservable for
two range and bearing measurements because the rotation around a specific unit
vector is undetermined. This can be overcome by using a third range and bearing
measurement and solving a system of linear equations for the rotation matrix.
Another method assumes that each robot is equipped with an IMU. An onboard IMU
makes the roll and pitch angles of each agent observable because the IMU is
able to measure the gravity vector which, when combined with vehicle motion,
allows the roll and pitch to be estimated accurately. Assuming the roll and
pitch estimates are accurate, they can be treated as known constants, making
the problem of estimating the relative rotation a two dimensional problem where
the relative heading can be solved for uniquely.

We solve for the relative pose using optimization techniques,
as opposed to the algebraic solutions in \cite{zhou2012determining}, to obtain
greater accuracy in our relative pose estimates. The problem can be
solved quickly given its small size. The optimization problem is
defined as follows
\begin{equation}
    T = \argmin_T \sum_{i=1}^n \lVert z - h(T, T_a, T_b)\rVert ^2_{\Sigma_k}
    \label{eq:rb_opt_prob}
\end{equation}
and $h(T, T_a, T_b)$ is defined as
\begin{equation}
    h(T, T_a, T_b) = \begin{pmatrix}
        \lVert dt \rVert \\
        \text{atan2}(dy, dx)
    \end{pmatrix}
\end{equation}
where $dt$ is the translation component of $(T_a^{-1}(T\,T_b))$ and $dx$ and
$dy$ are the components of $dt$. We solve the problem using Ceres
\cite{ceres-solver} assuming that $T_a$ and $T_b$ are known and constant.
We can then use the transform $T$ in the cycle constraints
defined in \cref{eq:cb_pgo}. A minimum of two measurements are needed to solve
\cref{eq:rb_opt_prob} but more can be used depending on the quality of the
measurements.


\section{Results}

In this section we provide extensive validation of the algorithms described
above. We outline the experiments used to evaluate all the algorithms
presented in \cref{sec:Methods} and provide a discussion of their implications.

\subsection{Incremental Algorithm Validation}
\label{subsec:inc_alg_val}

\begin{table*}[]
    \vspace{4mm}
    \centering
    \caption{Density for all of the benchmark datasets when all data has been used. \textbf{ICB} indicates using the incremental cycle basis algorithm and \textbf{MCB} indicates the minimum cycle basis algorithm.}
\begin{tabular}{|c|c|c|c|c|c|c|c|}
\hline
                     & \textbf{INTEL} & \textbf{MITb} & \textbf{M3500} & \textbf{City10k} & \textbf{Torus10000} & \textbf{Sphere2500} & \textbf{Kitti} \\ \hline
\textbf{ICB} & 2.62         & 1.34        & 2.79         & 3.25           & 3.31              & 2.48              & 1.40         \\ \hline
\textbf{MCB}         & 2.06         & 1.31        & 2.17         & 2.36           & 2.42              & 1.99              & 1.32         \\ \hline
\end{tabular}
\label{tab:density_table}
\end{table*}

\begin{table*}[]
    \centering
    \caption{Time required to update the cycle bases and run the optimization when for the following datasets. \textbf{ICB} is using the incremental cycle basis and \textbf{MCB} is using the minimum cycle basis.}
\begin{tabular}{|c|c|c|c|c|c|c|c|c|}
\hline
                                                                                            &              & \textbf{INTEL} & \textbf{MITb} & \textbf{M3500} & \textbf{City10k} & \textbf{Torus10000} & \textbf{Sphere2500} & \textbf{Kitti} \\ \hline
\multirow{2}{*}{\textbf{\begin{tabular}[c]{@{}c@{}}Cycle Basis\\ Update (ms)\end{tabular}}} & \textbf{ICB} & 0.057          & 0.076         & 0.14          & 0.59            & 0.81               & 0.13               & 0.17          \\ \cline{2-9}
                                                                                            & \textbf{MCB} & 174        & 49.0        & 1896        & 2189          & 2702              & 1030             & 1865        \\ \hline
\multirow{2}{*}{\textbf{\begin{tabular}[c]{@{}c@{}}Optimization\\ (ms)\end{tabular}}}       & \textbf{ICB} & 3.93           & 2.06         & 45.76         & 293          & 1668              & 234             & 6.56          \\ \cline{2-9}
                                                                                            & \textbf{MCB} & 2.90          & 1.65          & 26.7         & 184          & 1277             & 240             & 7.86          \\ \hline
\end{tabular}
    \label{tab:update_table}
    \vspace{-4mm}
\end{table*}

Since no experimental validation of the incremental cycle basis algorithm in
\cite{bai2020two} has been done, we first perform experiments to compare the
performance of the incremental and minimum cycle bases.

\subsubsection*{Cycle Basis Density}

Our first experiment was designed to compare the sparsity of the incremental
cycle basis with the MCB. This was deemed important since the use of the MCB
was done to maximize the sparsity of the problem. We define a metric to measure
the density as the following

\begin{equation}
    \rho = \frac{\sum_{\mathcal{C}_i \in \mathcal{C}} |\mathcal{C}_i|}{|\mathcal{E}|}.
    \label{eq:density_metric}
\end{equation}
The metric sums the number of edges in each cycle in the cycle basis, which
corresponds to the number of non-zero entries in the Jacobian, and normalizes it
by the number of edges in the graph. The MCB will have the minimum achievable
density. Our goal is to compare how well the incremental
cycle basis approximates the density of the MCB.
We provide the comparison on several standard benchmark datasets.

\begin{figure}[h]
    \centering
    \includegraphics[width=0.9\columnwidth]{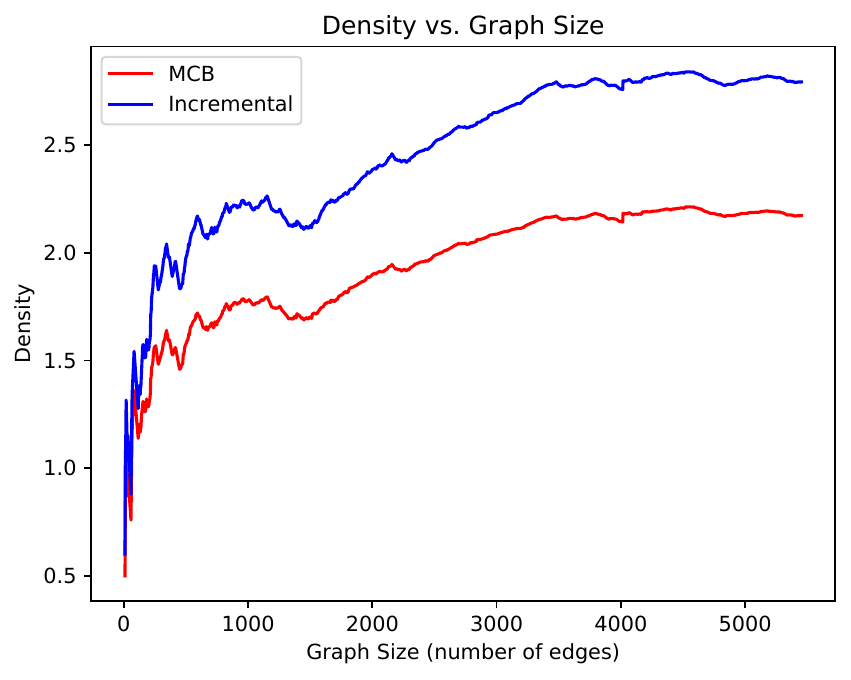}
    \caption{Visual comparison of the density of the incremental cycle basis and the MCB as the graph grows for the M3500 dataset.}
    \label{fig:density_figure}
\end{figure}

\Cref{fig:density_figure} contains a visualization of how the density of both
cycle bases change as the graph grows for the M3500 dataset \cite{teller2006fast} plotted against
the number of edges in the graph. \Cref{tab:density_table} contains the
density for the entirety of each dataset found in
\cite{carlone2014angular,carlone2015initialization}. As is
expected, the incremental algorithm produces a cycle basis that is denser than
the MCB. Observing the values in \cref{tab:density_table} shows that the
incremental algorithm produces a cycle basis that is about 1.3 times more dense
than the MCB. However, many of these datasets (M3500, City10k, Torus1000, and
Sphere2500) produce graphs that are more dense than those in many PGO
applications. The incremental algorithm is capable of producing a cycle basis
that has almost the same density of the MCB in sparser graphs, such as those in
the MITb and Kitti datasets.


\subsubsection*{Algorithm Complexity}

The complexity of both the incremental algorithm and the MCB algorithm in
\cite{bai2021sparse} were presented earlier. We provide additional validation
on the runtime of each algorithm. We test the time that it takes for both the
incremental algorithm and the MCB algorithm to update their respective cycle
bases. Since no incremental update to an MCB exists we report the time required
to recompute the cycle basis. We then record the update time to be plotted against the
number of edges in the graph.
We also record the time required to solve the optimization problem in
\cref{eq:cb_pgo} to determine the effect of the increased cycle basis density
on the optimization runtime. As
with the sparsity experiment, we provide a comparison across multiple benchmark
datasets.

\begin{figure} [tbh]
    \centering
    \includegraphics[width=0.9\columnwidth]{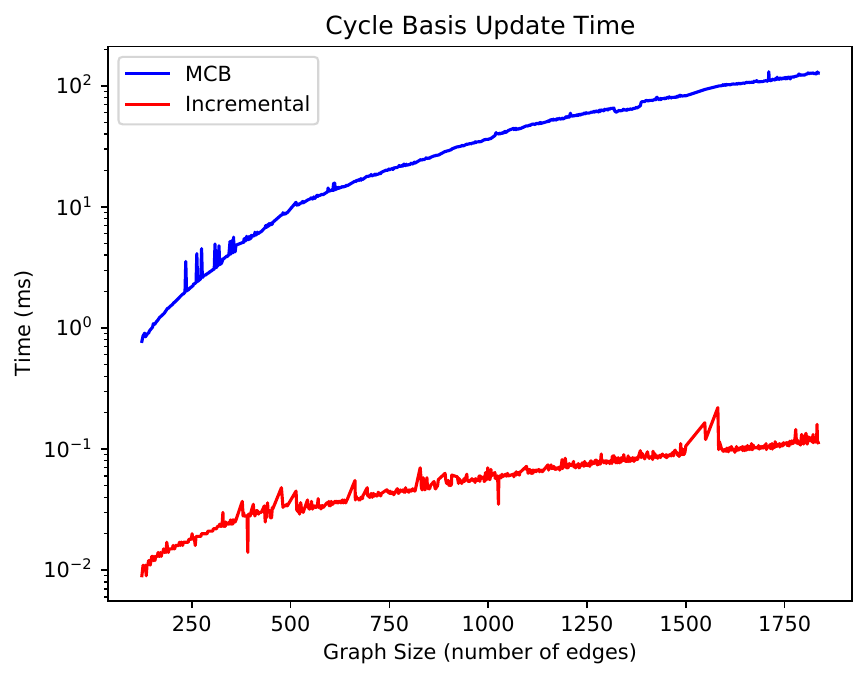}
    \caption{Time required to update the cycle bases for the incremental algorithm and MCB algorithm on the INTEL dataset. \textit{Note the log scale on the vertical axis.}}
    \label{fig:cb_update_figure}
\end{figure}

\begin{figure}[tbh]
    \centering
    \includegraphics[width=0.9\columnwidth]{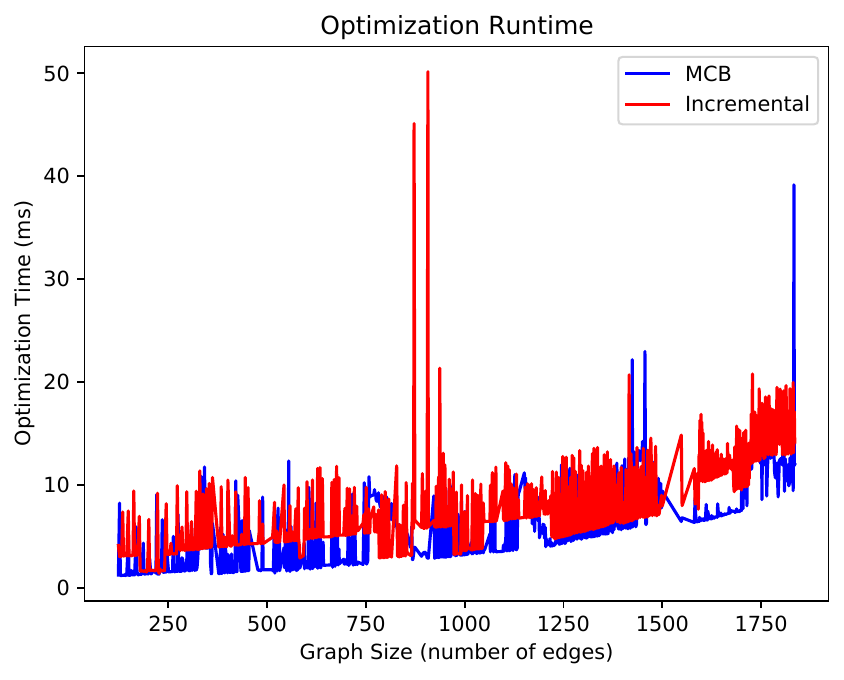}
    \caption{Time required for the optimization to converge on the INTEL dataset.}
    \label{fig:cb_optimization_figure}
\end{figure}

\Cref{fig:cb_update_figure} shows the time it takes to update the cycle bases
on the INTEL dataset and \cref{fig:cb_optimization_figure} shows the time it
takes for the optimization to reach a solution. Looking at
\cref{fig:cb_update_figure} and \cref{fig:cb_optimization_figure} we see
that the incremental algorithm can update a cycle basis in a time three orders
of magnitude faster than the MCB algorithm while having comparable optimization
speeds. Observing \cref{tab:update_table} shows a similar trend where the
incremental cycle basis update is several orders of magnitude faster than the
MCB calculation, but sacrifices comparatively little optimization speed despite
the increased cycle basis density.

\subsection{Multi-agent Cycle Basis Validation}


In this section we provide validation of the multi-agent incremental cycle
basis (MA-ICB) algorithm in \cref{alg:multi_agent_cb}. We first compare the density and
cycle basis update times of \cref{alg:multi_agent_cb} with those of the
algorithm in \cite{bai2020two}. We also compare the error statistics of
using the MCB and the cycle basis found in \cref{alg:multi_agent_cb} in both
simulation and hardware experiments.

\subsubsection{Performance Evaluation}
We validate \cref{alg:multi_agent_cb} using the same density and algorithm
complexity experiments presented in \cref{subsec:inc_alg_val} on the same
datasets but with modifications to simulate multi-agent scenarios. Each dataset
was split in half and the two halves were assumed to be collected
simultaneously. We recomputed the density, using \cref{eq:density_metric},
after every inter-agent loop closure. We show the benefit of being able to
parallelize \cref{alg:multi_agent_cb} by measuring the time to update the cycle
basis for both algorithms and comparing them.
Additionally, to show the value of recomputing
inter-agent loops, we show the density after recomputing the inter-agent loops
at the end of the simulation.

\begin{table*}[]
    \vspace{4mm}
    \centering
    \caption{Comparison of the density and average update times of \cref{alg:multi_agent_cb} and the incremental algorithm in \cite{bai2020two}.}
    \label{tab:macb_alg_val}
\begin{tabular}{|c|c|c|c|c|c|c|c|}
\hline
                                                                                             &               & \textbf{INTEL} & \textbf{M3500} & \textbf{City 10k} & \textbf{Torus10000} & \textbf{Sphere2500} & \textbf{Kitti} \\ \hline
\multirow{2}{*}{\textbf{\begin{tabular}[c]{@{}c@{}}Average Update\\ Time (ms)\end{tabular}}} & \textbf{ICB}  & 0.026          & 0.122          & 0.40              & 0.43                & 0.13                & 0.39           \\ \cline{2-8}
                                                                                             & \textbf{MA-ICB} & 0.039          & 0.082          & 0.29              & 0.27                & 0.052               & 0.19           \\ \hline
\multirow{2}{*}{\textbf{Density}}                                                            & \textbf{ICB}  & 2.16           & 2.46           & 2.77              & 2.79                & 2.00                & 1.38           \\ \cline{2-8}
                                                                                             & \textbf{MA-ICB} & 6.02           & 5.41           & 11.43             & 8.62                & 2.00                & 1.84           \\ \hline
\textbf{\begin{tabular}[c]{@{}c@{}}Recomputed \\ Density\end{tabular}}                       & \textbf{MA-ICB} & 2.52           & 2.62           & 3.32              & 3.38                & 2.00                & 1.40           \\ \hline
\end{tabular}
\end{table*}

We present our results in \cref{tab:macb_alg_val} where we present the density
at the end of the simulation, the average time to update the cycle basis, and
the density of the cycle basis using \cref{alg:multi_agent_cb} when the
inter-agent loops have been recomputed at the end of the simulation. As can be
seen \cref{alg:multi_agent_cb} improves the update time by around 30 percent on
most datasets. We note that there is significant variation of the density of
the cycle basis produced by \cref{alg:multi_agent_cb} and this is primarily a
result of the edge ordering we utilize in \cref{alg:multi_agent_cb}. However,
the density can be siginifcantly reduced by recomputing inter-agent loops as
the graphs for each individual agent become more developed. In our experiment
we recomputed all inter-agent loops and this usually required tens of
milliseconds to complete. Further examination of the original loops produced by
\cref{alg:multi_agent_cb} shows that a minority of the loops contribute to
significant portion of the density. A more targeted recomputation strategy
could reduce the time required while significantly decreasing the density by
identifying and recomputing the longest inter-agent loops.

\subsubsection{Simulation Experiment}
Our simulation experiment  was done to compare the error in both the edges and
the trajectory with the ground truth data produced in the simulator. Our
simulator generated 10 agents that operated in a Manhattan-world-like
trajectory. Each robot has 500 odometry edges and 100 intra-vehicle loop
closures. Between any given pair of robots there are 50 inter-vehicle loop
closure measurements that are generated randomly. Noise was added to both the
odometry and loop closure measurements using the noise characteristics of the
M3500 dataset in \cite{carlone2014angular}. The optimization problem in
\cref{eq:cb_pgo} was solved using the cycle constraints from both the MCB and
the cycle basis defined in \cref{alg:multi_agent_cb}. After solving the
optimization, we compute the root-mean-squared error (RMSE) of the edges,
defined in \cref{eq:rmse}, and the average trajectory error (ATE)
of the poses, defined in \cref{eq:ate}, for all agents.
The $\text{trans}(\cdot)$ operator in \cref{eq:ate} indicates the translation
component of the rigid-body transformation.

\begin{equation}
    \text{RMSE} = \sqrt{\frac{\sum_{i=1}^N \lvert \lvert \text{Log}(dT_i^{-1}\hat{dT}_i) \rvert \rvert ^2}{N}}
    \label{eq:rmse}
\end{equation}

\begin{equation}
    \text{ATE} = \sqrt{\frac{\sum_{i=1}^N \lvert \lvert \text{trans}(T_i^{-1}\hat{T}_i) \rvert \rvert ^2}{N}}
    \label{eq:ate}
\end{equation}

We report the RMSE of the translational components and the
rotational components of the edges separately, and our results appear in
\cref{tab:ma_sim_err_results}. Observing \cref{tab:ma_sim_err_results} we find
that the two cycle basis algorithms produce nearly identical results, where the
small differences can be attributed to different stopping points that satisfied
the stopping criteria of the optimization. The time required to compute all the
required cycles between graphs using \cref{alg:multi_agent_cb} was $118$ ms
while the time required using the MCB algorithm was $4170$ ms, showing that we
can achieve similar accuracy while using only a small fraction of the time to
compute cycle constraints.

\begin{table}[b]
    \caption{Results comparing the error of the MCB and the MA-ICB algorithm in \cref{alg:multi_agent_cb}.}
    \label{tab:ma_sim_err_results}
\begin{tabular}{|c|c|c|c|}
\hline
\textbf{Algorithm} & \textbf{Rot. RMSE (deg)} & \textbf{Trans. RMSE (m)} & \textbf{ATE (m)} \\ \hline
\textbf{MCB}       & 5.74                   & 0.39                   & 0.55           \\ \hline
\textbf{MA-ICB}    & 5.74                   & 0.39                   & 0.56           \\ \hline
\end{tabular}
\end{table}

\begin{table*}[t]
    \vspace{4mm}
    \centering
    \caption{Error results for both the single and multi-agent hardware experiments.}
    \label{tab:hardware_exp_err_res}
\begin{tabular}{|c|c|c|c|c|}
\hline
\textbf{Experiment}                    & \textbf{Algorithm} & \textbf{Rot. RMSE (deg)} & \textbf{Trans. RMSE(m)} & \textbf{ATE (m)} \\ \hline
\multirow{2}{*}{\textbf{Single agent}} & \textbf{MCB}       & 2.65                 & 0.070                & 0.19         \\ \cline{2-5}
                                       & \textbf{ICB}       & 2.90                  & 0.068                 & 0.22          \\ \hline
\multirow{2}{*}{\textbf{Multi-agent}}  & \textbf{MCB}       & 0.63                 & 0.024                & 0.052         \\ \cline{2-5}
                                       & \textbf{MA-ICB}       &  0.63                 & 0.023                 & 0.033          \\ \hline
\end{tabular}
\vspace{-6mm}
\end{table*}

\subsubsection{Hardware Experiment}
In this section, we describe the experimental validation of the algorithms presented in \cref{alg:multi_agent_cb} and in \cite{bai2020two}. The experiments are performed on Turtlebots carrying an Intel NUC processor on board running Ubuntu 18.04. The Robot Operating System is used to facilitate message passing between the sensors and agents. A motion capture system is used to generate edges every 0.5 meters or a change of 15 degrees in yaw  as a stand-in for vision based alternatives like \cite{koch2020relative}. We test Bai's incremental algorithm in a single-agent scenario and our algorithm in a multi-agent experiment.

\begin{figure}
    \centering
    \includegraphics[width=0.9\columnwidth]{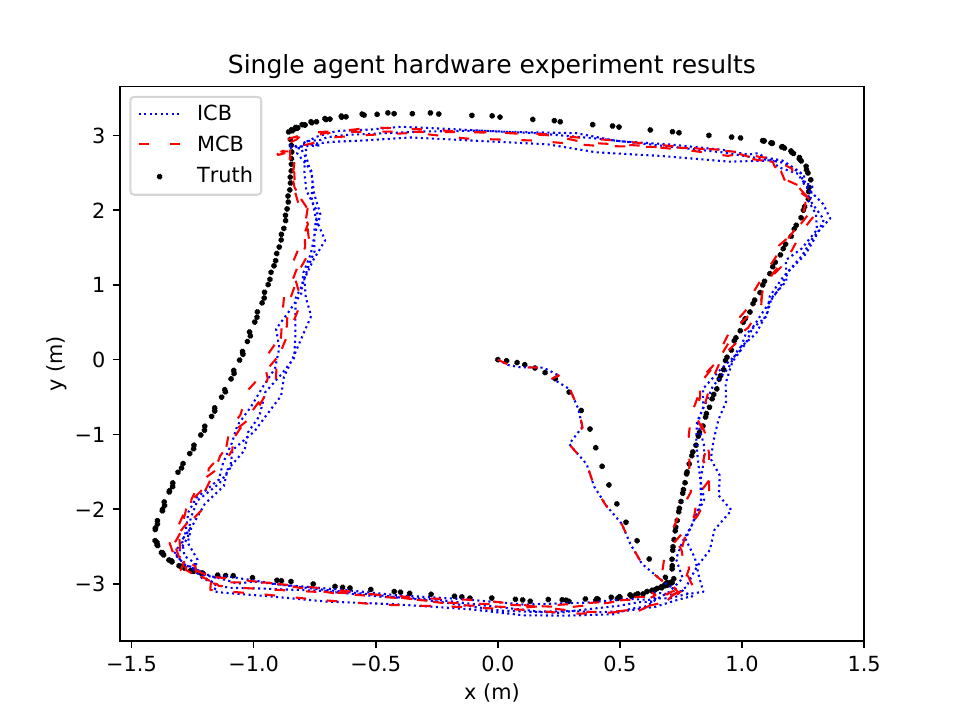}
    \caption{Results of the single agent hardware experiment using a cycle basis described by Bai \cite{bai2020two}.}
    \label{fig:single_agent_hardware_result}
\end{figure}

In the single-agent case, the ground robot follows a rectangular trajectory with feedback from the motion capture system to generate loop closures when the Turtlebot returned to a location it had previously been. The results of this experiment are shown in \cref{fig:single_agent_hardware_result}. The average RMS error for this experiment can be found in \cref{tab:hardware_exp_err_res}. The algorithm took $1.46$ ms to optimize over 210 edges using the MCB and $1.87$ ms using the incremental cycle basis.

\begin{figure}
    \centering
    \includegraphics[width=0.9\columnwidth]{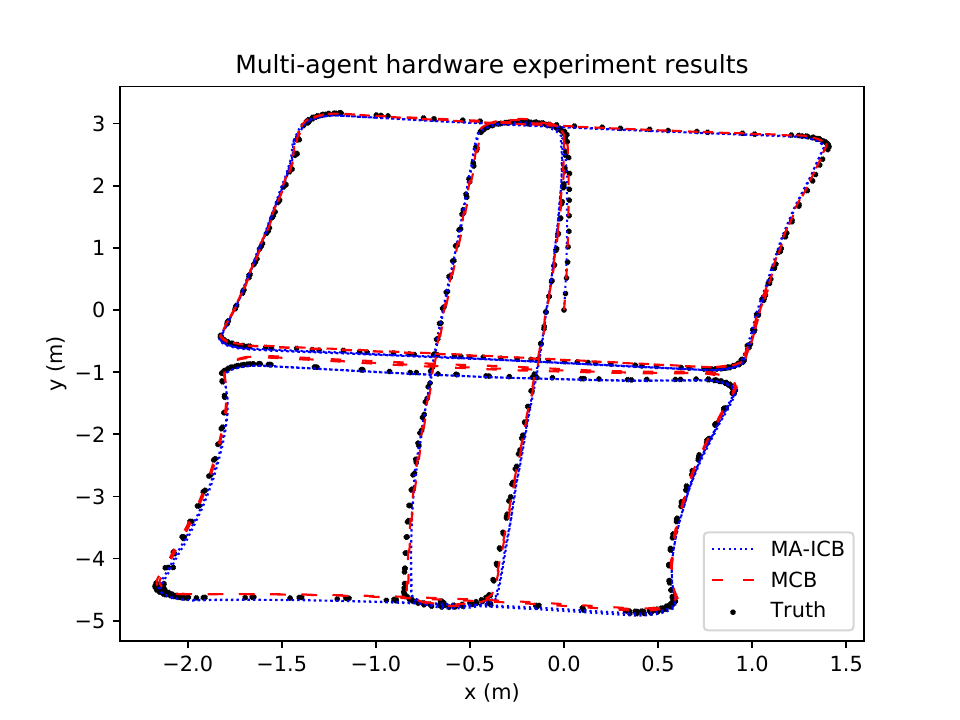}
    \caption{Results of the multi-agent experiment using a cycle basis generated by \cref{alg:multi_agent_cb}.}
    \label{fig:ma_hardware_result}
\end{figure}

In the multi-agent scenario, three Turtlebots are used and follow rectangular trajectories. Each agent has its own processing unit and generates edges as described in the single-agent experiment. Periodically, agents will communicate with each other. During a communication, agents will share the data they have collected and obtain a relative pose between them. In this experiment, agents 0, 1 and 2 have 209, 217 and 193 edges, respectively, and 20 different communication events evenly distributed among pairs of robots. The results of this run are shown in \cref{fig:ma_hardware_result} and show the results from the perspective of agent 0. The RMS error for this run when compared to the motion capture truth reference is  shown in \cref{tab:hardware_exp_err_res}, along with the ATE. We also compared the time to update the cycle bases on the last optimization. Updating the MCB took $49.6$ ms, while updating the the cycle bases between all pairs of robots using \cref{alg:multi_agent_cb} required $0.66$ ms.

Observing the results in \cref{fig:single_agent_hardware_result},
\cref{fig:ma_hardware_result}, and those in \cref{tab:hardware_exp_err_res}
show that choice of cycle basis has little impact on the accuracy obtained by
the optimization. In combination with the results from
\cref{subsec:inc_alg_val}, we show that we can obtain this accuracy and
significantly lower the computational time, especially for large graphs, by
using algorithms to approximate a MCB instead of exactly computing one.

\subsection{Low-DOF Measurements}

\begin{figure}
    \centering
    \includegraphics[width=0.9\columnwidth]{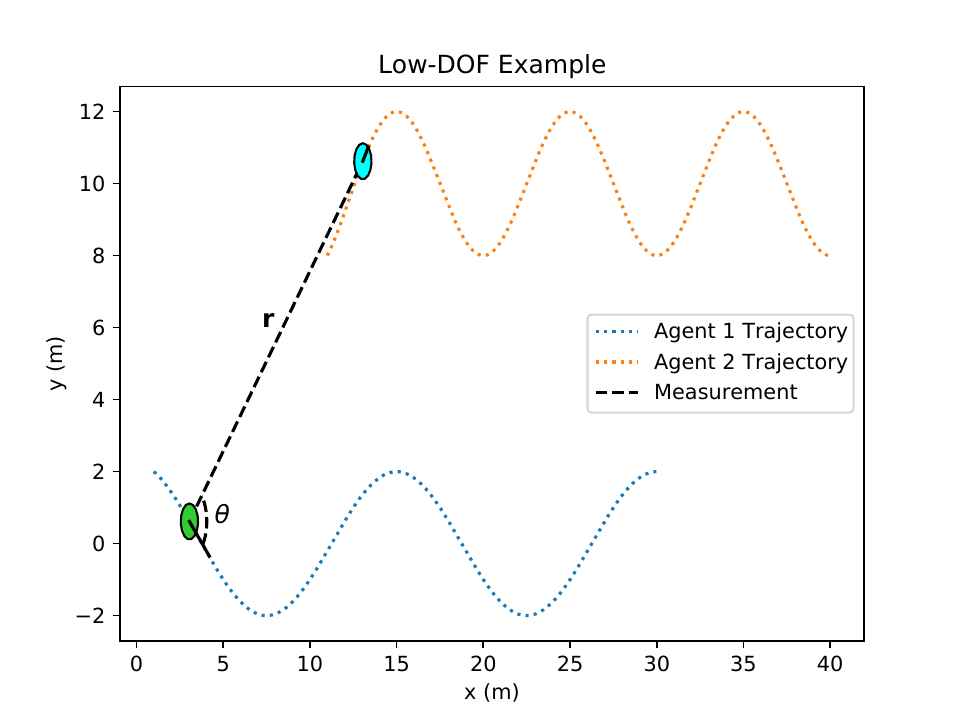}
    \caption{Example trajectories for the two agents. Note that each agent travels in the direction it is facing. The range measurement $r=\lvert \lvert \mathbf{r} \rvert \rvert$ and the bearing measurement is $\theta=\text{atan2}(\mathbf{r}_y, \mathbf{r}_x) - \theta_1$.}
    \label{fig:rb_sample_traj}
\end{figure}

This experiment was designed to validate the use of low-DOF
measurements in the relative parameterization of PGO. In this experiment,
two agents were simulated in a planar environment traveling along sine
waves of different frequencies and phase offsets. Odometry was
simulated by taking two poses along the trajectory at times between 0.5 and 2.5
seconds apart, computing the transformation between them,
and adding noise. This approach created between 75 and 95 poses per agent
along a 120 second long simulation. 30 range and bearing measurements
were computed at random intervals along the trajectories as shown in
\cref{fig:rb_sample_traj}. These measurements were used to create
inter-vehicle loop closures by solving \cref{eq:rb_opt_prob} using pairs of
sequential measurements. This resulted in 15 relative-pose measurements to be
used in the cycle-based PGO problem. We ran
30 different trials to record and average the error in the relative
localization between the agents. The noise characteristics used in the
experiment are described below.
\begin{equation*}
    \begin{aligned}
        \Sigma_{\text{odom}} &= \text{diag}(0.01\text{m}^2, 0.01\text{m}^2, 0.001\text{rad}^2) \\
        \Sigma_{\text{meas}} &= \text{diag}(0.1\text{m}^2, 0.01\text{rad}^2)
    \end{aligned}
\end{equation*}

Since the purpose of multi-agent PGO is for the agents to
localize relative to one another, we evaluate the error in the relative pose
estimates at the start and the end of the trajectory. We compute the average
and standard deviation of the error across the different trials and report the
results in \cref{tab:rb_relative_pose_err_table}. We compare the results of our
method with the results produced by GTSAM \cite{dellaert2012factor} when range
and bearing factors are used to model inter-agent measurements.

As can be seen, we obtain accurate relative-pose estimates with the average
error between the two agents being less than a meter in the translational
components and less than three degrees in the relative orientation showing that
low-DOF measurements can effectively be used in the cycle-based PGO problem
described in \cref{eq:cb_pgo}. Additionally, our method obtains similar
translational accuracy when compared with GTSAM but suffers both lower relative
heading accuracy and larger variance on the relative heading error. This
degredation in the relative heading is expected as we do not recompute the
relative pose factors by solving \cref{eq:rb_opt_prob} using updated estimates
for $T_a$ and $T_b$. The accuracy in the relative heading factors, $T$, are
dependent on the quality of  the odometry of each robot and the errors in the
measurements.



\begin{table*}[t]
    \vspace{4mm}
    \centering
    \caption{Relative pose error results for the low-DOF measurement experiment.}
    \label{tab:rb_relative_pose_err_table}
\begin{tabular}{|c|c|cc|cc|}
\hline
\textbf{Time}                     & \textbf{Variable}              & \multicolumn{2}{c|}{\textbf{Mean Relative Pose Error}} & \multicolumn{2}{c|}{\textbf{Std. Deviation}} \\ \hline
\multicolumn{1}{|l|}{}   & \multicolumn{1}{l|}{} & \multicolumn{1}{c|}{\textbf{CBPGO}}       & \textbf{GTSAM}      & \multicolumn{1}{c|}{\textbf{CBPGO}}  & \textbf{GTSAM} \\ \hline
\multirow{3}{*}{0 (s)}                    & x (m)                 & \multicolumn{1}{c|}{-0.0017}     & -0.12      & \multicolumn{1}{c|}{0.84}   & 1.39  \\ \cline{2-6}
                         & y (m)                 & \multicolumn{1}{c|}{0.019}       & 0.024      & \multicolumn{1}{c|}{0.94}   & 1.40  \\ \cline{2-6}
                         & $\theta$ (deg)        & \multicolumn{1}{c|}{-0.14}       & 0.14       & \multicolumn{1}{c|}{6.69}   & 0.86  \\ \hline
\multirow{3}{*}{120 (s)} & x (m)                 & \multicolumn{1}{c|}{0.010}       & -0.022     & \multicolumn{1}{c|}{1.37}   & 1.51  \\ \cline{2-6}
                         & y(m)                  & \multicolumn{1}{c|}{-0.78}       & -0.29      & \multicolumn{1}{c|}{1.38}   & 2.41  \\ \cline{2-6}
                         & $\theta$ (deg)        & \multicolumn{1}{c|}{-3.02}      & -0.68      & \multicolumn{1}{c|}{13.6}   & 1.00  \\ \hline
\end{tabular}
\vspace{-2.5mm}
\end{table*}

\section{Conclusion}

In summary, we have validated a proposed but untested algorithm to
incrementally compute a sparse cycle basis and shown that the incremental cycle
basis closely approximates the MCB. Additionally, we presented and
validated an algorithm to approximate the MCB of the sparse
connection of two disjoint graphs and demonstrated its ability to run in
real-time PGO scenarios and give accurate estimates of the edges in each
robot's graph. Lastly, we have introduced a methodology to utilize
low degree of freedom measurements in a relative PGO framework and
demonstrated that the method produces accurate relative-pose measurements
between the different agents. These improvements take steps toward allowing
relative PGO to be efficiently used in multi-agent scenarios where obtaining a
relative-pose measurement between agents can be difficult.




\section*{Acknowledgement}
We would like to acknowledge Paul Buzaud and Jared Paquet for their help setting up the hardware for the experiments in this paper.

\bibliographystyle{style/IEEEtran}
\bibliography{references/IEEEabrv, references/strings-short, references/library}

\end{document}